# Deep Learning for Weather Forecasting: A CNN-LSTM Hybrid Model for Predicting Historical Temperature Data

**Yuhao Gong[1,2], Yuchen Zhang[1,3], Fei Wang[1,4,*], Chi-Han Lee[2,5]**

[1] School of Information and Artificial Intelligence, Nanchang Institute of Science and Technology, Nanchang, China
[2] School of professional Studies, Applied Analytics, Columbia University In the City of New York, NYC, USA.

[2] 1713029162@qq.com
[3] 2627556529@qq.com
[4] wangfei2002@163.com
[5] cl4069@columbia.edu
*corresponding author

**Abstract.** As global climate change intensifies, accurate weather forecasting has become increasingly important, affecting agriculture, energy management, environmental protection, and daily life. This study introduces a hybrid model combining Convolutional Neural Networks (CNNs) and Long Short-Term Memory (LSTM) networks to predict historical temperature data. CNNs are utilized for spatial feature extraction, while LSTMs handle temporal dependencies, resulting in significantly improved prediction accuracy and stability. By using Mean Absolute Error (MAE) as the loss function, the model demonstrates excellent performance in processing complex meteorological data, addressing challenges such as missing data and high-dimensionality. The results show a strong alignment between the prediction curve and test data, validating the model's potential in climate prediction. This study offers valuable insights for fields such as agriculture, energy management, and urban planning, and lays the groundwork for future applications in weather forecasting under the context of global climate change.

## 1. Introduction

As global climate change intensifies, the importance of accurate weather forecasting has escalated, impacting agriculture, energy management, environmental protection, and daily life. The precision of weather predictions is crucial for socio-economic development and the overall quality of life. Traditional forecasting methods, relying on physical models and statistical approaches, often struggle to handle the complexities of nonlinear and time-varying features, particularly under the uncertainties posed by climate change.

Recent advancements in deep learning have emerged as promising solutions for improving weather prediction accuracy. Techniques such as Convolutional Neural Networks (CNNs) effectively capture spatial patterns, while Recurrent Neural Networks (RNNs) and Long Short-Term Memory networks (LSTMs) excel in processing time series data, effectively managing time dependencies in meteorological variables. Despite notable progress in predicting rainfall and temperature trends, single-model approaches often falter when faced with high-dimensional, multi-scale weather data. This has led researchers to explore hybrid models, particularly the CNN-LSTM combination, to enhance prediction performance.

This study specifically focuses on analyzing and predicting historical temperature data, utilizing a CNN-LSTM hybrid model. The research involves a comprehensive workflow that includes data preprocessing, exploratory analysis, and model training, demonstrating the hybrid model's strong performance in prediction accuracy and stability. By effectively addressing challenges such as missing data and the complexities of high-dimensional meteorological information, this research showcases the potential of deep learning technologies in weather forecasting.

## 2. Related Work

Time series forecasting has traditionally relied on models like Autoregressive Moving Average (ARMA) and Autoregressive Integrated Moving Average (ARIMA), known for their effectiveness with linear and stationary data in fields such as economic forecasting and energy consumption. However, these methods are limited by their linear assumptions, struggling with nonlinear and time-varying datasets **[1]**. In response, Recurrent Neural Networks (RNNs) and Long Short-Term Memory (LSTM) networks emerged, capturing long-term dependencies in sequential data, and proving useful in areas like speech recognition. Yet, LSTMs can encounter challenges with intricate local patterns and overfitting, especially in noisy environments **[2]**.

To enhance performance, hybrid models combining Convolutional Neural Networks (CNNs) and LSTMs have been developed, capitalizing on CNNs' ability to extract local features while utilizing LSTMs for temporal modeling. This CNN-LSTM architecture has shown promise in various applications, including precipitation forecasting, though it introduces computational complexity and overfitting risks, particularly with high-dimensional data **[3]**.

Recent advancements have increasingly incorporated deep learning techniques into weather forecasting, especially in regions with complex climates. These models have significantly improved the accuracy of temperature and weather predictions. Notable studies include Ranjan et al., who introduced a hybrid neural network for predicting traffic congestion, and Zhang et al., who employed a CNN-LSTM model for long-term global temperature forecasting, demonstrating enhanced prediction accuracy **[4-5]**. Huang and Kuo applied CNN-LSTM for PM2.5 concentration forecasting, achieving superior results compared to other machine learning methods **[6]**. Wang et al. explored CNN-LSTM for short-term storm surge prediction, finding that the hybrid model outperformed traditional approaches like Support Vector Regression **[7]**.

Overall, the integration of CNNs and LSTMs has marked a significant advancement in time series forecasting, particularly in meteorological applications, by effectively managing complex data patterns and improving prediction accuracy. These developments offer valuable insights for future research and practical applications in various domains, including climate change mitigation and urban planning **[8]**.

## 3. Predicting Temperature Model

### 3.1. Data

This dataset primarily consists of eight fields: Region, Country, State, City, Month, Day, Year, and AvgTemperature. The dataset covers multiple regions, with North America accounting for 54%, Europe for 13%, and other regions for 33%. This distribution indicates that the amount of data from North America is significantly higher than that from other regions. The dataset does not specify individual countries in detail, but it can be inferred that the majority of data originates from North America and Europe. This diversity may provide a comparative basis for climate characteristics across different regions.

The data is primarily concentrated in the state of Texas, which represents 4% of the total. Data from other states or regions is classified as "Other," accounting for 46%. This may reflect a regional preference in data collection, which should be considered in the analysis.

In terms of cities, data has been collected from a total of 321 cities and regions. The Month variable records the observation months, which helps in analyzing seasonal climate changes and temperature trends. Data from different months can reveal inter-annual temperature variation patterns. The Day variable records the specific observation dates, providing a foundation for analyzing daily temperature fluctuations. This variable, when combined with the Month variable, aids in understanding temperature changes within the season.

The Year variable records the observation years, with a substantial amount of data from 1983 to 2020 (approximately 2,905,887 entries). This time span provides an important temporal dimension for studying climate change and its impacts. The AvgTemperature variable records the average temperature for each day, with data distributed across multiple ranges, showing a significant number of records for both low and high temperatures. This variable is central to climate analysis, revealing overall trends and patterns of temperature variation.

Each variable holds unique importance in data analysis, contributing to a deeper understanding of temperature changes and the factors behind them.

### 3.2. Variable introduction

The main data used in this study consists of daily average temperature values from several major cities. These temperature data are typically provided by meteorological departments, environmental monitoring agencies, and relevant government authorities to ensure accuracy and reliability. The data spans different seasons and years, reflecting the changing trends of urban climate. To enhance the representativeness of the analysis, cities with varying climate characteristics and geographical locations were selected to fully capture the diversity of temperature changes. This data not only provides essential historical temperature information but also serves as a solid foundation for model training and validation.

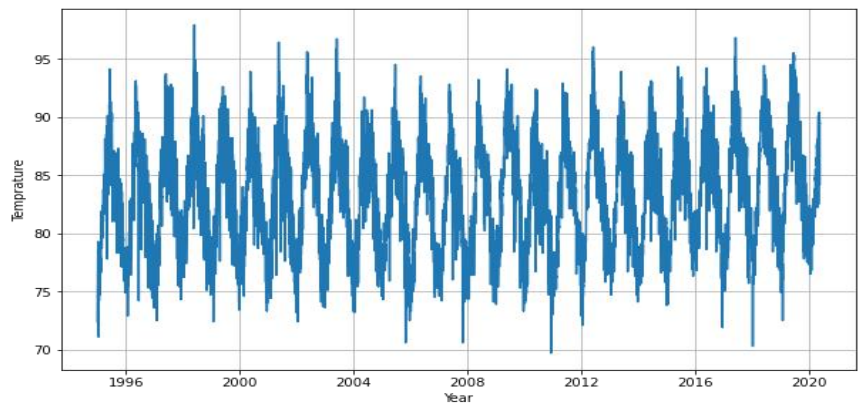

**Figure 1.** Structure of CNN-LSTM Model

### 3.3. CNN-LSTM Model

Table 1 summarizes the architecture of the model, detailing each layer's output shape and associated parameters. The Input_layer has a shape of (None, 60, 1), accepting sequences of length 60 with a single feature, and contains no parameters. The first Conv1D layer, with 60 filters, produces an output shape of (None, 56, 60) and introduces 360 parameters for convolutional operations. Two LSTM layers, each with 60 units, maintain an output shape of (None, 60, 60) and consist of 24,840 parameters each, enabling the capture of long-term dependencies in the data. The first Dense layer reduces the output to 30 units, resulting in a shape of (None, 30) with 1,830 parameters. The second Dense layer further decreases this to 10 units, producing an output shape of (None, 10) and incorporating 310 parameters. The final Dense layer outputs a single prediction with 11 parameters.

Lastly, the Lambda layer applies a scaling transformation to the output, maintaining the shape (None, 1) and containing no additional parameters. Overall, this table provides a comprehensive overview of the model's structure and how each layer contributes to processing and predicting time series data **[9-10]**.

Table 1. All the layers with specific parameters.

| Layer | Out Shape | Param |
|---|---|---|
| Input_layer | (None, 60, 1) | 0 |
| Conv1D (filters=60) | (None, 56, 60) | 360 |
| LSTM (units=60) | (None, 60, 60) | 24840 |
| LSTM (units=60) | (None, 60, 60) | 24840 |
| Dense (units=30) | (None, 30) | 1830 |
| Dense (units=10) | (None, 10) | 310 |
| Dense (units=1) | (None, 1) | 11 |
| Lambda | (None, 1) | 0 |

The first Conv1D layer utilizes 60 convolutional filters, each with a kernel size of 5. By applying the ReLU activation function, this layer introduces non-linearity into the model, enabling it to efficiently extract local features from the input sequence. The input consists of 60 time steps, each containing one feature, making this layer essential for identifying initial patterns in the data. The second Conv1D layer continues the feature extraction process, maintaining 60 filters to refine and capture higher-level features. The ReLU activation function ensures that the model can recognize more complex patterns within the time series data.

Following the convolutional layers, the model includes two LSTM layers, each comprising 60 units. These layers are crucial for capturing long-term dependencies within the input sequence, as they process the output from the previous layers. Both LSTM layers return sequences, preserving the temporal information necessary for subsequent processing. The LSTM architecture enables the model to understand the intricate relationships between data points over time, enhancing its predictive capabilities.

The model then transitions to a series of Dense layers, starting with one that outputs 30 units. This layer reduces the output dimensionality while maintaining the essential features extracted from the previous layers. The next Dense layer further decreases the output to 10 units, allowing for more compact representation of the learned features. Finally, a Dense layer with a single output unit is employed to generate the final prediction, effectively summarizing the information learned throughout the model.

The last layer in the model is a Lambda layer, which applies a scaling transformation to the final output. This layer ensures that the predicted values are adjusted appropriately, maintaining a shape of (None, 1) with no additional parameters. By incorporating this transformation, the model enhances its ability to provide accurate and meaningful predictions based on the processed data.

## 4. Results

The selection of the loss function is a critical factor that significantly influences the prediction accuracy of the model. In this study, we have chosen to utilize Mean Absolute Error (MAE) as the primary loss function for evaluating the model's performance. MAE quantifies the average magnitude of the errors in a set of predictions by calculating the absolute differences between predicted values and actual outcomes. One of the key advantages of MAE is that it does not take into account the direction of these errors; instead, it focuses solely on their size. This characteristic makes MAE particularly suitable for tasks where the objective is to minimize the overall prediction errors, regardless of whether those errors are positive or negative.

Furthermore, MAE provides a clear and intuitive interpretation of the model's performance. By representing the average error in the same units as the predicted values, it allows for easy comparison and understanding of the model's accuracy. For example, if the predictions are in degrees Celsius, the MAE will also be expressed in degrees Celsius, making it straightforward for practitioners to gauge how closely the model's predictions align with actual temperature readings. This clarity is invaluable in practical applications, where stakeholders often seek to understand the implications of model performance in tangible terms. Overall, the use of MAE as the loss function not only enhances the reliability of the model's predictions but also aids in communicating its effectiveness in a meaningful way.

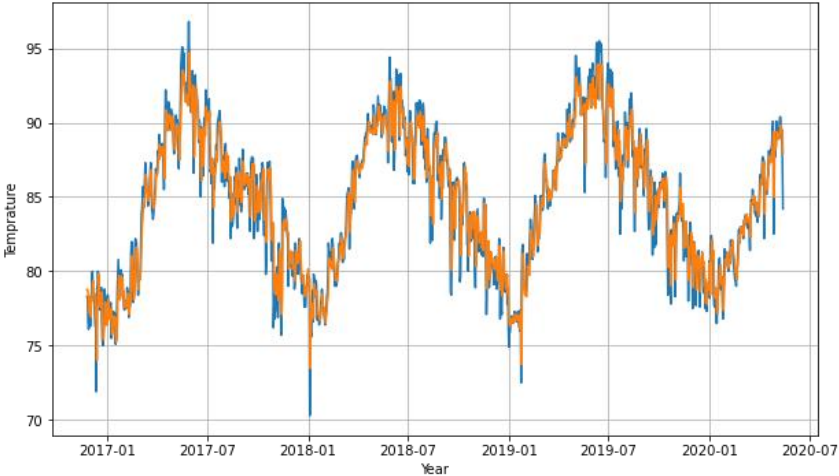

**Figure 3.** The prediction curve and test curve.

This study examines the model's convergence by plotting the loss function's variation curve during the training process. The results indicate that the Mean Absolute Error (MAE) experiences a rapid decline from over 20 in the early stages of training, quickly approaching a value near 1. This sharp decrease illustrates the model's effective learning mechanism, as it swiftly adapts to the underlying patterns present in the training data. Following this initial phase, the MAE stabilizes, ultimately reaching a value close to 0.90. This stability is crucial, as it not only signifies that the model has successfully learned the significant features and trends within the dataset but also reflects an enhanced prediction accuracy. The consistent reduction in the loss function throughout training highlights the model's robustness and reliability in capturing the complexities of the temperature data, thereby reinforcing its effectiveness in weather forecasting. Overall, the convergence observed in the loss function illustrates the model's capacity to generalize well to unseen data, which is a critical aspect of its performance.

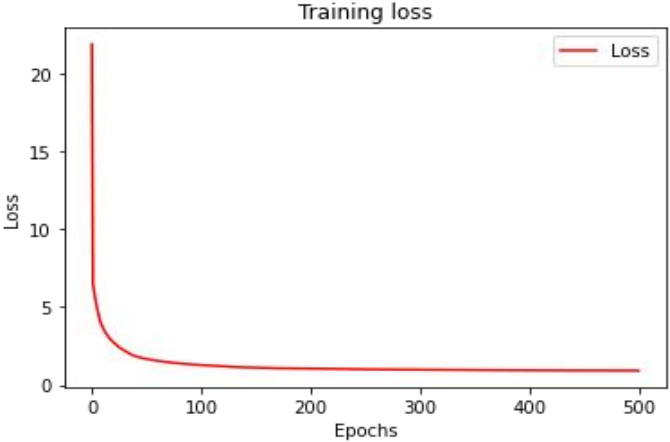

**Figure 4.** Training Loss curve.

The results demonstrate that the CNN-LSTM model performs well on these metrics, with high prediction accuracy and stability. The prediction curve's high similarity to the test curve further supports the model's effectiveness in temperature prediction.

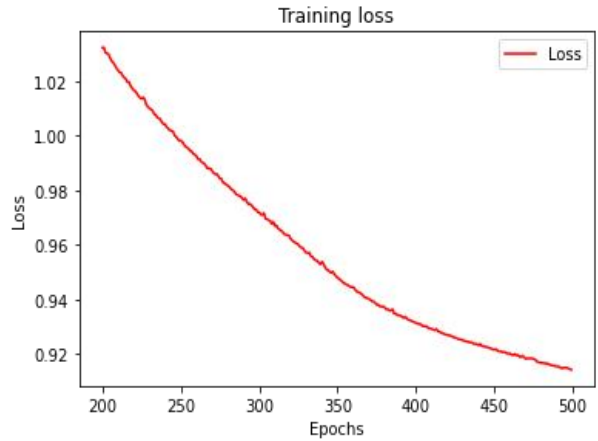

**Figure 5.** Training Loss - Zoomed In curve.

The performance comparison of various models for predicting historical temperature data, as outlined in Table 2, highlights the advantages of utilizing more sophisticated approaches in deep learning.

Linear Regression serves as the baseline model, achieving a variance of 0.682, an $R^2$ score of 0.623, and a Mean Absolute Error (MAE) of 2.125. While it provides a foundational understanding of the data, its performance indicates limitations in capturing the complexities of temperature patterns.

Convolutional Neural Network (CNN) improves upon linear regression, with a variance of 0.756 and an $R^2$ score of 0.791. The MAE decreases to 1.536, demonstrating the effectiveness of CNNs in capturing spatial features of the temperature data. This model shows a clear advancement in accuracy, although it still lacks the temporal processing capabilities necessary for comprehensive time series forecasting.

**Table 2.** Comparison of Model Performance

| Model | Variance | R2 Score | MAE |
|---|---|---|---|
| Linear Regression | 0.682 | 0.623 | 2.125 |
| CNN | 0.756 | 0.791 | 1.536 |
| LSTM | 0.873 | 0.890 | 1.018 |
| CNN- LSTM | 0.927 | 0.901 | 0.901 |

Long Short-Term Memory (LSTM) networks further enhance prediction performance, achieving a variance of 0.873 and an $R^2$ score of 0.890. The MAE drops significantly to 1.018, underscoring the LSTM's strength in modeling sequential data and capturing long-term dependencies. This model illustrates a marked improvement over both linear regression and CNN, highlighting its suitability for time series forecasting.

CNN-LSTM Hybrid Model demonstrates the highest performance metrics, with a variance of 0.927 and an $R^2$ score of 0.901. The MAE is further reduced to 0.901, confirming the model's effectiveness in addressing both spatial and temporal aspects of the temperature data. This hybrid approach capitalizes on the strengths of both CNN and LSTM architectures, resulting in superior prediction accuracy and stability.

Overall, the analysis reveals that the CNN-LSTM hybrid model outperforms the other approaches, offering a robust solution for forecasting complex meteorological data. This underscores the potential of integrating deep learning techniques to enhance the accuracy of weather predictions, which can be beneficial for various applications in climate-related fields.

## 5. Conclusions

This study implemented a CNN-LSTM hybrid model to predict historical temperature data demonstrating the efficacy of deep learning techniques in weather forecasting. The hybrid model effectively combined the strengths of Convolutional Neural Networks (CNNs) for spatial feature extraction and Long Short-Term Memory (LSTM) networks for capturing temporal dependencies, resulting in high prediction accuracy and stability. The use of Mean Absolute Error (MAE) as the loss function provided a clear and interpretable evaluation of the model's performance.

The results indicated that the CNN-LSTM model excelled at handling complex meteorological data, addressing key challenges such as missing values and high-dimensionality. The strong alignment between the prediction curve and the actual test data highlighted the model's reliability and its potential for broader applications in climate prediction.

These findings offer valuable insights for sectors such as agriculture, energy management, and urban planning. Future research could explore the application of this model to other regions with similar weather variability and extend its use to predict additional weather-related phenomena, further enhancing the capabilities of meteorological forecasting in the context of global climate change.